\documentclass{article} 
\usepackage{iclr2027_conference,times}

\usepackage{amsmath,amsfonts,bm}

\def\eqref#1{equation~\ref{#1}}

\def\1{\bm{1}}

\DeclareMathAlphabet{\mathsfit}{\encodingdefault}{\sfdefault}{m}{sl}
\SetMathAlphabet{\mathsfit}{bold}{\encodingdefault}{\sfdefault}{bx}{n}

\usepackage{hyperref}
\usepackage{url}
\usepackage{graphicx}
\usepackage{booktabs}

\usepackage{booktabs}
\usepackage{longtable}
\usepackage{makecell}
\usepackage{multirow}
\usepackage{marvosym}

\title{$\Delta$WAM: Distilling Action Tangent Fields into World Action Models}

\arxivpreprint

\author{Ke Wu$^{1,2}$ \quad Hanwen Huang$^{1,2}$ \quad Bo Gu$^{1,2}$ \quad Kaizhao Zhang$^{1,2}$ \quad Xiangting Meng$^{3}$ \\ 
\textbf{Yupeng Zheng}$^{4}$ \quad \textbf{Zijun Xu}$^{1}$ \quad \textbf{Jieru Zhao}$^{5}$ \quad \textbf{Wenchao Ding}$^{1,2}$\thanks{Corresponding author.} \\[0.15in]
$^{1}$Fudan University \quad $^{2}$TARS Robotics \quad $^{3}$ShanghaiTech University \\
$^{4}$Institute of Automation, Chinese Academy of Sciences \quad
$^{5}$Shanghai Jiao Tong University
}

\begin{document}

\maketitle

\begin{figure}[h]
    \centering
    \vspace{-1.1cm}
    \includegraphics[width=\linewidth]{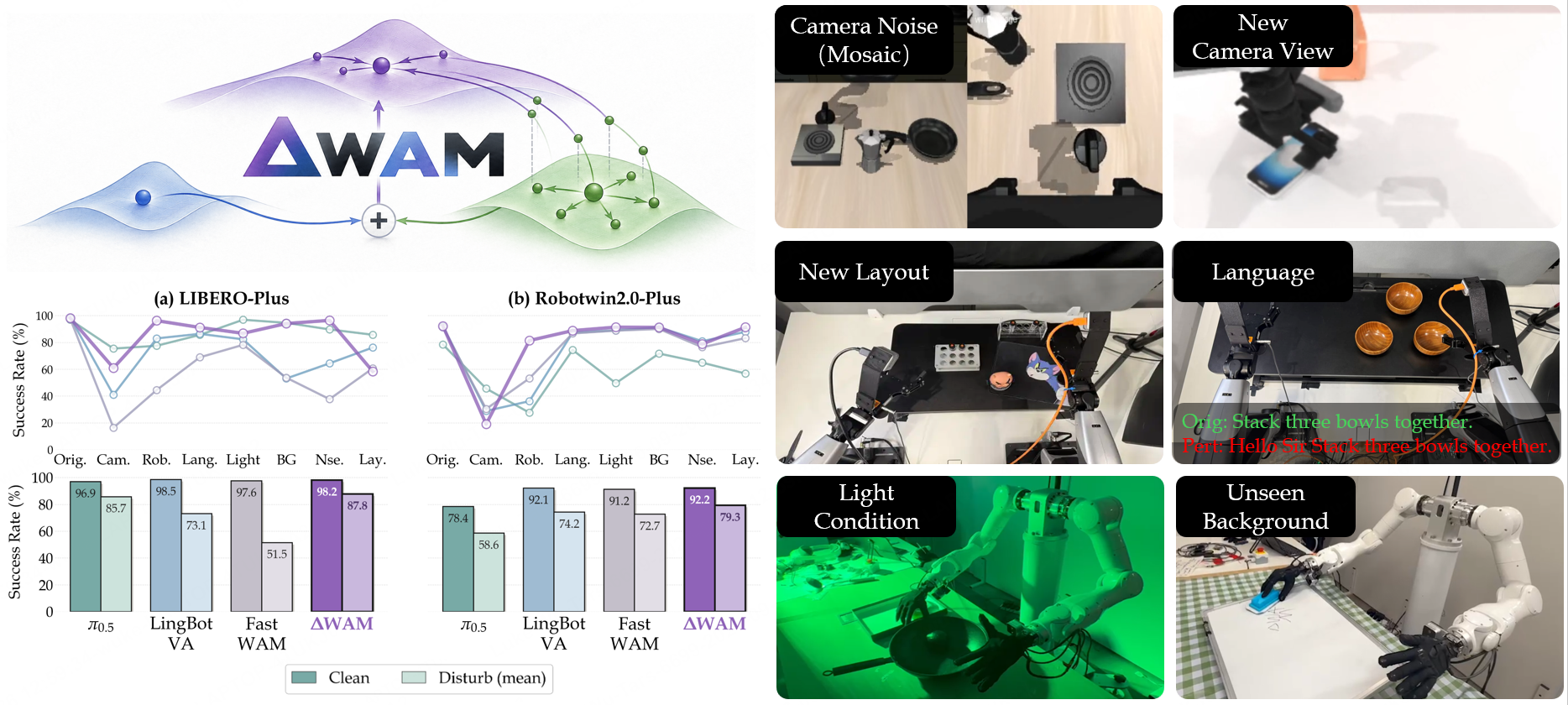}
    \caption{\textbf{$\Delta$WAM under distribution shift}. Left: success rate by perturbation type on LIBERO-Plus and RoboTwin 2.0-Plus, with the average over perturbations shown for each method. Right: the corresponding shifts, camera noise, novel viewpoints and layouts, language, lighting, and unseen backgrounds.}
    \vspace{-0.2cm}
\end{figure}

\begin{abstract}
World Action Models (WAM) improve robot policies by augmenting sparse action supervision with dense future prediction. However, much of the predictable future is dominated by appearance and scene persistence rather than action-dependent dynamics. We observe that several recent WAM designs, including optical flow, motion-centric representations, and latent actions, can be understood from a common perspective in which world supervision becomes more efficient as it contains a higher proportion of action-relevant variation. Based on this insight, we introduce Action Tangent Fields, which reformulate world supervision through a local Taylor expansion of how actions induce changes in future dynamics. We represent future dynamics in Residual-VAE space, where the future latent remains recoverable from the current latent and its residual, and use a strong action-conditioned world model (ACWM) to probe the local correspondence between action variations and residual-world variations. This local first-order structure is distilled into the WAM to guide its denoising supervision toward dynamics that are more tightly coupled to action, rather than merely predictable from appearance. Across LIBERO-Plus, RoboTwin, and RoboTwin2.0-Plus, our method consistently improves robustness to lighting, background, camera, layout, and other environmental perturbations. Despite using no large-scale embodied pretraining, it achieves stronger robustness under several distribution shifts than pretrained policies. We further distill multi-step VideoDiT denoising into a single step for efficient inference. Our results suggest that effective WAM supervision should remain information-rich while concentrating its predictive capacity on the directions along which actions change the future.
\end{abstract}

\section{Introduction}
Behavior cloning labels each demonstrated observation with an action. World Action Models (WAMs) also predict the future video, giving dense supervision for object motion, contact, and state changes that a single action label cannot describe. A video target, however, rewards predictability rather than control relevance. Backgrounds, camera views, and untouched objects are already determined by the current frame and persist as the robot acts, so a model that reproduces them reduces its loss substantially while learning little about the consequences of its own actions. The value of world supervision therefore depends less on how much of the future it covers than on how much of that future is action-dependent.

Recent WAM designs change the target itself. Optical-flow and motion-centric targets keep only what moves, and latent-action targets compress the future into a compact code. Both reduce the supervision spent on static appearance, which is why they help, but both rely on motion as a proxy for action-dependence: a lighting change or a moving distractor alters the future with no contribution from the robot, and dropping appearance removes cues that contact-rich manipulation needs. A change of target also cannot reveal which directions of the future respond to the action, because each demonstration records one action and the single future that followed it. It is an isolated point of the action--dynamics relation, and its local neighborhood is never observed.

Our first step therefore keeps the target information-rich and changes its coordinate instead. Given the current frame, we subtract a broadcast reference derived from it, so the residual describes how the scene departs from the observed frame rather than restating it. Reference and residual recover the future VAE latent exactly, so no information is lost, and because the reference depends only on the current frame, subtracting it leaves the response to an action perturbation unchanged.

The second step supplies the missing neighborhood through a probe. For a fixed scene, an embodiment-specific action-conditioned world model (ACWM) is queried at small, temporally smooth perturbations of the expert action. The resulting residual futures differ mainly along directions the ACWM treats as action-sensitive, and their normalized differences span an Action Tangent Field (ATF), a low-dimensional estimate of those directions. Post-training keeps every element of the demonstrated residual in the loss and adds weight along the ATF, which makes the world loss anisotropic, ordinary world post-training is the isotropic special case. The probe sets the weighting, never the targets.

We first mid-train the Wan VideoDiT on a modest amount of robot manipulation video to adapt its prediction space to the residual latent space, then post-train a Fast-WAM-style policy with cached ATFs. Caching keeps the ACWM out of policy training and inference. Extensive experiments show that supervising dynamics-relevant world representations enables our method to generalize to visual perturbations substantially better than competing methods.

Our contributions are as follows.
\begin{itemize}
    \item We introduce an information-preserving Residual-VAE target that shifts world prediction toward scene dynamics while retaining exact recovery of the future video.

    \item Motivated by a first-order Taylor expansion of the action-to-future dynamics map, we introduce Action Tangent Fields to estimate locally action-responsive directions and emphasize them during denoising supervision of the real demonstrated future.

    \item We obtain efficient world features with one VideoDiT evaluation at inference. Aggregate success improves from 51.5\% to 87.8\% over Fast-WAM on LIBERO-Plus and from 72.7\% to 79.3\% on RoboTwin 2.0-Plus.
\end{itemize}

\section{Related Work}

\subsection{World Action Models}

World Action Models (WAMs) complement sparse action labels with future prediction, encouraging policies to retain object states, interactions, and temporal dynamics. Predict-then-act methods such as GE-Act generate compact visual futures and recover controls through inverse dynamics, while GE-Act~2.0 scales this design with heterogeneous interaction data~\citep{geact}. Joint models instead couple world and action generation: LingBot-VA autoregressively interleaves both outputs, whereas Cosmos Policy represents future observations, actions, and values as latent frames of a video diffusion model~\citep{lingbotva,cosmospolicy}.

A complementary line treats prediction primarily as representation supervision. Fast-WAM and Light-WAM retain video co-training but omit explicit generation at deployment~\citep{fastwam,lightwam}, whereas Efficient-WAM and Faster-WAM preserve future latents while reducing generation or video--action interaction costs~\citep{efficientwam,fasterwam}. The prediction target nevertheless determines how model capacity is allocated: full video latents preserve rich details but are dominated by static appearance and other variations weakly related to control, whereas flow, residual, and motion-centric targets emphasize change but may discard useful cues. In contrast, Action Tangent Fields retain an information-rich residual future space and selectively emphasize locally action-responsive directions.

\subsection{Action-Conditioned World Models}

Action-conditioned world models (ACWMs) predict visual consequences under specified controls. Recent systems broaden the control interface: Cosmos~3 jointly models language, images, video, audio, and actions~\citep{cosmos3}, Hydra-0 represents actions as pixel-space flow for cross-embodiment transfer~\citep{hydra0}, and Pelican-Sim combines a unified numerical action space with rendered action videos~\citep{pelicansim}. These methods condition generation through native action tokens, visual motion, and action-rendered observations, respectively.

Another line improves simulation reliability. ABot-PhysWorld uses physics-aware data and preference optimization to suppress implausible interactions~\citep{abotphysworld}. Ctrl-World combines multi-view prediction, frame-level action conditioning, and pose-aware memory for policy-in-the-loop rollouts~\citep{ctrlworld}, while WALL-SS uses coarse-to-fine autoregression and compressed memory for long streaming generation~\citep{wallss}. WorldSync addresses failures on off-expert commands through broader action coverage, dynamics grounding, and intervention-effect alignment~\citep{worldsync}. Unlike these approaches, we neither plan with generated rollouts nor use synthetic futures as supervision targets. Instead, we perturb demonstrated actions and use the ACWM as a local differential probe in Residual-VAE space, transferring the resulting first-order action--world geometry while retaining real demonstrations as prediction targets.




\section{Method}
\label{sec:method}

\begin{figure*}[t]
    \centering
    \includegraphics[width=0.95\textwidth]{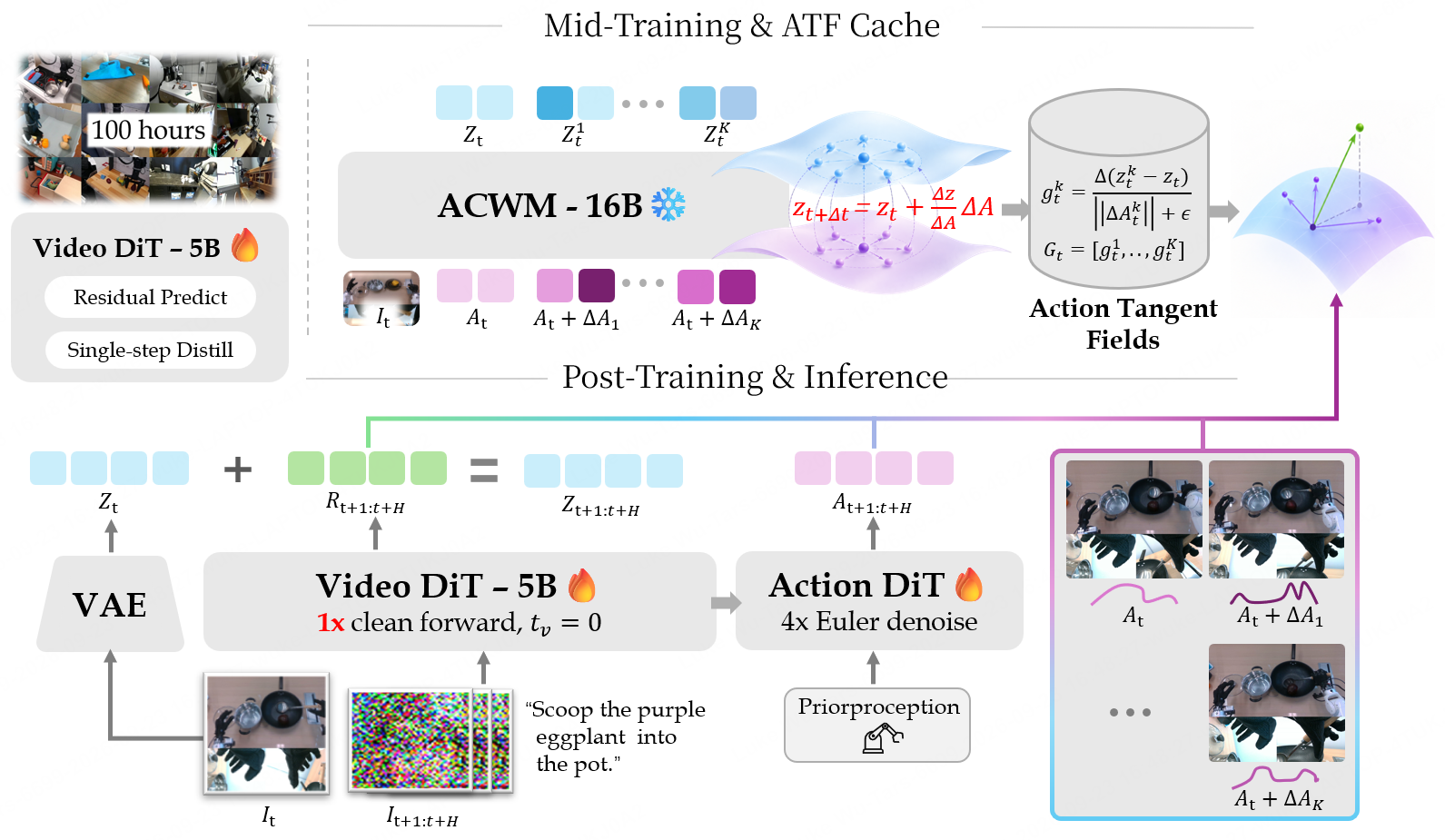}
    \caption{\textbf{Overview of $\Delta$WAM}. A Wan VideoDiT-5B is mid-trained to predict the Residual-VAE future $R_{t+1:t+H}$. An embodiment-specific ACWM is then queried at $K$ perturbed actions, and its normalized response differences are orthonormalized into a cached basis $G_t$ of locally action-sensitive directions. Post-training keeps the real residual as the target and penalizes its projection onto $G_t$, one VideoDiT pass at $t_v=0$ conditions four ActionDiT Euler steps at inference.}
    \label{fig:delta-wam-pipeline}
\end{figure*}

We introduce $\Delta$WAM to train a world-action policy on real future video while transferring local action sensitivity from an action-conditioned world model (ACWM). Figure~\ref{fig:delta-wam-pipeline} outlines the method. We first adapt Wan to predict future video latents relative to the current frame~\citep{wan2025wan}. We then perturb demonstrated actions and query an embodiment-specific ACWM~\citep{cosmos3} to extract action tangent fields (ATFs), which identify locally action-responsive directions. Finally, we post-train a Fast-WAM-style VideoDiT--ActionDiT policy~\citep{fastwam} on real action--future pairs, giving additional weight to prediction errors along those directions. At inference, one VideoDiT pass provides the world features used to generate an action chunk.

\subsection{Residual Video Mid-Training}
\label{sec:residual-midtraining}

The pretrained Wan VideoDiT predicts absolute future VAE latents, whereas our policy predicts future latents relative to the observed frame. Although the two representations contain the same information given that frame, they are different prediction targets. We therefore mid-train Wan before policy post-training to adapt its prediction space from absolute VAE latents to Residual-VAE latents. The known first frame carries persistent scene content, leaving the VideoDiT to predict changes such as robot motion, object displacement, and contact.

Let $E$ and $D$ denote the frozen Wan VAE encoder and decoder. We encode the observed first frame and the real future clip, then broadcast the first-frame code across the future latent positions:
\begin{equation}
    \begin{aligned}
    Z_t&=E(I_t),&
    Z_{t+1:t+H}&=E(I_{t+1:t+H}),\\
    B_t&=\operatorname{broadcast}(Z_t),&
    R_{t+1:t+H}&=Z_{t+1:t+H}-B_t.
    \end{aligned}
    \label{eq:delta-residual-target}
\end{equation}
Here $\operatorname{broadcast}$ aligns the first-frame code with the shape of $Z_{t+1:t+H}$. The anchor $B_t$ is known at prediction time, only the residual $R_{t+1:t+H}$ must be predicted. This change of coordinates loses no information relative to the original Wan latent:
\begin{equation}
    Z_{t+1:t+H}=B_t+R_{t+1:t+H},\qquad
    D(B_t+R_{t+1:t+H})=D(Z_{t+1:t+H}).
    \label{eq:delta-residual-recovery}
\end{equation}
Thus the first-frame anchor plus the future residual recovers exactly the absolute latent, and the Wan decoder produces the same video it would have produced from that latent. The claim is about preserving the VAE representation, not exact recovery of the original RGB pixels. The subtraction centers prediction on departures from the observed scene while retaining changes in robot pose, object state, and contact. Because $B_t$ is fixed when $I_t$ is fixed, subtracting it does not change the local response to an action perturbation: $\Delta R_{t+1:t+H}=\Delta Z_{t+1:t+H}$. This lets ATF identify action-related directions in the same information-rich latent space used by the VideoDiT.

For this prediction-space adaptation, we select 100 hours of robotics video from the DROID~\cite{khazatsky2024droid} and BridgeV2~\cite{walke2023bridgedata} portions of the Cosmos training corpus. We initialize the VideoDiT from pretrained Wan and train it to predict residuals using the same flow-matching parameterization. For video flow time $t_v$, Gaussian noise $\epsilon_R$, and the Wan interpolation coefficients $\alpha_{t_v}$ and $\sigma_{t_v}$, the noisy residual and its velocity target are
\begin{equation}
    x_{t,t_v}^{R}=\alpha_{t_v} R_{t+1:t+H}+\sigma_{t_v}\epsilon_R,
    \qquad
    u_{t,t_v}^{R,\star}=\dot\alpha_{t_v} R_{t+1:t+H}+\dot\sigma_{t_v}\epsilon_R.
    \label{eq:delta-residual-flow}
\end{equation}
The residual VideoDiT $F_{\theta_{\mathrm{mid}}}$ minimizes
\begin{equation}
    \mathcal L_{\mathrm{mid}}=
    \mathbb E_{t,t_v,\epsilon_R}
    \left[\left\|F_{\theta_{\mathrm{mid}}}
    (x_{t,t_v}^{R},t_v,I_t,l_t)-u_{t,t_v}^{R,\star}\right\|_2^2\right].
    \label{eq:delta-midtrain-loss}
\end{equation}
This stage aligns the pretrained VideoDiT with the Residual-VAE target used in policy post-training, it requires video but no action labels. The residual captures what changed, while the next stage uses a local action neighborhood to identify which changes respond to action.

\subsection{Action Tangent Fields from Local Dynamics}
\label{sec:atf-cache}

The demonstrated residual $R_{t+1:t+H}$ gives one realization of future dynamics under $A_t$. To ask which parts of these dynamics depend on the action, imagine the unknown map $f_t(A)$ from a nearby action chunk to the vectorized future residual at the same current state. Where this map is differentiable, a small action change produces
\begin{equation}
    f_t(A_t+\Delta A)-f_t(A_t)
    \approx J_t\operatorname{vec}(\Delta A),
    \qquad
    J_t=\left.\frac{\partial f_t}{\partial\operatorname{vec}(A)}\right|_{A_t}.
    \label{eq:delta-desired-tangent}
\end{equation}
The directions in the output of $J_t$ are the possible local changes in video dynamics induced by action. A single demonstrated future cannot reveal them. We therefore adapt a 16B ACWM~\citet{cosmos3} and posttrain them separately for each robot embodiment and use it to sample nearby futures. Let $b(t)$ identify the embodiment and $\Phi_{b(t)}$ its frozen ACWM. The ACWM provides a neighborhood of dynamics, not replacement targets for the policy.

We sample $K=8$ small, temporally smooth perturbations $\Delta A_t^k$ in normalized action coordinates and set $A_t^k=A_t+\Delta A_t^k$. Normalization makes perturbation magnitudes comparable across action dimensions, while random directions probe the neighborhood without selecting a preferred coordinate axis. For the reference and perturbed actions we hold the observation, instruction, proprioception, and generation conditions $\xi_t$ fixed:
\begin{equation}
    \widetilde I_t^0=\Phi_{b(t)}(I_t,l_t,p_t,A_t,\xi_t),
    \qquad
    \widetilde I_t^k=\Phi_{b(t)}(I_t,l_t,p_t,A_t^k,\xi_t).
    \label{eq:delta-acwm-rollouts}
\end{equation}
We sample frames at the policy's prediction times and encode the generated futures with the same frozen VAE. In Figure~\ref{fig:delta-wam-pipeline}, $z_t=E(\widetilde I_t^0)$ is the ACWM future under the expert action and $z_t^k=E(\widetilde I_t^k)$ is its future under perturbation $k$, lowercase $z$ distinguishes these generated futures from the observed latent $Z_t$. Their difference removes the common first-frame anchor:
\begin{equation}
    \begin{aligned}
    \Delta R_t^k
    &=(z_t^k-B_t)-(z_t-B_t)=z_t^k-z_t,\\
    \operatorname{vec}(\Delta R_t^k)
    &\approx \widetilde J_t\operatorname{vec}(\Delta A_t^k).
    \end{aligned}
    \label{eq:delta-world-response}
\end{equation}
Here $\widetilde J_t$ is the ACWM's local action--world Jacobian. The finite difference therefore estimates a direction in which \emph{future dynamics} change under a nearby action. It approximates the teacher's local response, not the unknown environment Jacobian $J_t$ in ~\eqref{eq:delta-desired-tangent}. Matched generation conditions help isolate action effects from sampling variation.

We divide each response by perturbation magnitude so that its scale reflects a directional sensitivity, then retain the output directions rather than attempting to estimate all entries of $\widetilde J_t$:
\begin{equation}
    g_t^k=
    \frac{\operatorname{vec}(\Delta R_t^k)}
    {\|\Delta A_t^k\|_2+\varepsilon},
    \qquad
    G_t=[g_t^1,\ldots,g_t^K]\in\mathbb R^{D_R\times K},
    \qquad
    Q_t=\operatorname{orth}(G_t).
    \label{eq:delta-tangent-basis}
\end{equation}
Here $D_R$ is the number of elements in one future residual. If $U_t$ collects the normalized perturbation directions, the finite-difference relation gives $G_t\approx\widetilde J_t U_t$. Thus the columns of $Q_t\in\mathbb R^{D_R\times r_t}$, where $r_t\leq K$ and $Q_t^\top Q_t=I_{r_t}$, span only the ACWM output directions reached by the eight sampled actions. The point is to locate a small, action-responsive neighborhood of the dynamics target, not to recover the full Jacobian. We cache $Q_t$ and its rank with the training window, the ACWM is not used during policy training or inference.

\subsection{ATF-Guided World-Action Post-Training}
\label{sec:atf-posttraining}

The first stage gives the VideoDiT a dense target for real future dynamics, the second marks directions in that target that vary across a local action neighborhood. We combine them during Fast-WAM-style post-training. The policy VideoDiT is initialized from the residual-mid-trained Wan backbone and supplies world features to an ActionDiT. Its world branch is conditioned on the observed scene and instruction, not on a candidate action, so we use $Q_t$ to weight errors in predicted dynamics rather than asking the policy to reproduce an action Jacobian.

During post-training, the VideoDiT uses the same single-pass computation as at inference. At fixed video time $t_v=0$, it processes the known latent reference $B_t$ and produces world tokens $h_t^W$. With the flow-output sign convention of our implementation, its output head $O_\theta$ gives a clean residual estimate:
\begin{equation}
    h_t^W=F_\theta(B_t,l_t,t_v=0),
    \qquad
    \widehat R_{t+1:t+H}=-O_\theta(h_t^W)_{\mathrm{future}}.
    \label{eq:delta-single-pass-world}
\end{equation}
The target is always the residual of the \emph{real} demonstrated future, and the ordinary loss covers every latent element:
\begin{equation}
    \mathcal L_R=
    \mathbb E_t\left[\frac{1}{D_R}
    \|\operatorname{vec}(\widehat R_{t+1:t+H}-R_{t+1:t+H})\|_2^2\right].
    \label{eq:delta-posttrain-world-loss}
\end{equation}
For action flow time $\tau$ and Gaussian noise $\epsilon_A$, let $A_{t,\tau}=(1-\tau)A_t+\tau\epsilon_A$. The ActionDiT predicts the velocity of this interpolation from the noisy action, proprioception, instruction, and world tokens:
\begin{equation}
    \mathcal L_A=
    \mathbb E_{t,\tau,\epsilon_A}
    \left[\left\|U_\phi(A_{t,\tau},\tau,p_t,l_t,h_t^W)
    -(\epsilon_A-A_t)\right\|_2^2\right].
    \label{eq:delta-posttrain-action-loss}
\end{equation}

The action and residual losses learn from the observed action--future pair. To make the world loss sensitive to its local action neighborhood, define the real-future dynamics error $e_t=\operatorname{vec}(\widehat R_{t+1:t+H}-R_{t+1:t+H})$ and add a penalty on its projection onto $Q_t$:
\begin{equation}
    \mathcal L_{\mathrm{ATF}}=
    \mathbb E_t\left[
    \frac{1}{r_t}\|Q_t^\top e_t\|_2^2
    \right],
    \label{eq:delta-atf-projection-loss}
\end{equation}
with a zero contribution when $r_t=0$. Since $Q_t$ has orthonormal columns, this is the squared error along the sampled action-responsive directions. For each window with $r_t>0$, the two world losses can be written as
\begin{equation}
    \begin{aligned}
    \frac{\lambda_R}{D_R}\|e_t\|_2^2+
    \frac{\lambda_T}{r_t}\|Q_t^\top e_t\|_2^2
    =e_t^\top\!
    \left(\frac{\lambda_R}{D_R}I_{D_R}
    +\frac{\lambda_T}{r_t}Q_tQ_t^\top\right)e_t.
    \end{aligned}
    \label{eq:delta-atf-metric}
\end{equation}
Every direction retains the ordinary residual weight, while directions in $\operatorname{span}(Q_t)$ receive additional weight. The full future latent is therefore still learned, but errors aligned with local action effects matter more.

The complete post-training objective is
\begin{equation}
    \mathcal L_{\mathrm{post}}=
    \lambda_A\mathcal L_A+
    \lambda_R\mathcal L_R+
    \lambda_T\mathcal L_{\mathrm{ATF}}.
    \label{eq:delta-posttrain-total}
\end{equation}
The VideoDiT still learns the dynamics of the real demonstrated future. The ATF term uses nearby, ACWM-generated outcomes only to identify which residual errors should matter more for action. Thus pointwise action and video targets are enriched by local action-neighborhood structure without training on synthetic futures.

\subsection{Single-Pass World Conditioning at Inference}
\label{sec:single-pass-inference}

At a replanning step, the policy receives the latest image $I_t$, instruction $l_t$, and proprioceptive state $p_t$. It encodes $I_t$ to $Z_t$, forms the broadcast reference $B_t$, and evaluates the VideoDiT once at $t_v=0$, exactly as in ~\eqref{eq:delta-single-pass-world}. The resulting world tokens $h_t^W$ summarize the observed scene and the dynamics learned during residual and ATF-guided post-training. Action generation consumes these tokens directly: it does not need to decode $\widehat R_{t+1:t+H}$ into video. Because the world branch is independent of the candidate action chunk, $h_t^W$ is computed once and reused throughout action sampling.

We sample an initial action chunk $a_t^{(0)}\sim\mathcal N(0,I)$ and integrate the learned action flow from noise time $\tau_0=1$ to data time $\tau_4=0$ using four descending Euler steps:
\begin{equation}
    a_t^{(j+1)}
    =a_t^{(j)}+(\tau_{j+1}-\tau_j)
    U_\phi\!\left(a_t^{(j)},\tau_j,p_t,l_t,h_t^W\right),
    \qquad j=0,\ldots,3.
    \label{eq:delta-inference-action-euler}
\end{equation}
The negative time increment follows the action-flow parameterization in ~\eqref{eq:delta-posttrain-action-loss}, where $\tau=1$ denotes noise and $\tau=0$ denotes the expert action. Only the noisy action chunk and its flow time change between Euler steps, the image encoding and VideoDiT features remain fixed. The final state $\widehat A_{t+1:t+H}=a_t^{(4)}$ is the predicted action chunk.


\section{Experiments}

We evaluate $\Delta$WAM on three simulation benchmarks and two real robot platforms. LIBERO-Plus and RoboTwin 2.0-Plus measure each perturbation dimension separately, which shows where action-tangent supervision helps and where it does not. RoboTwin, RoboTwin 2.0-Plus, and Tianji involve bimanual manipulation, and the real-world tasks add sustained contact and a 20-DoF dexterous hand action space. The real-world platform detail is provided in the Appendix. We then ablate the number of VideoDiT denoising steps and the number of action perturbations.

\subsection{LIBERO-Plus}




LIBERO-Plus~\cite{libero-plus} extends LIBERO~\cite{liu2023libero} with controlled shifts along seven dimensions: camera viewpoint, robot initialization, language, lighting, background, observation noise, and object layout. We fine-tune on the official demonstrations for 10 epochs and evaluate one checkpoint under every condition against vision-language-action policies (VLAs), VLA-plus-world-model methods, and WAMs. We obtain the best aggregate success rate of \textbf{87.8\%}, 2.1 points above the strongest baseline $\pi_{0.5}$. Our 98.2\% original-setting rate is also close to the best reported result of 98.6\%, indicating that the aggregate robustness gain does not come at a substantial cost to in-distribution performance, and does not require the large-scale embodied pretraining that several competing VLAs inherit. Total is the pooled success rate across all perturbation instances (total successes divided by total evaluated instances, see Appendix A.2 and Table 4).

\begin{table}[t]
    \centering
    \caption{
        \textbf{LIBERO-Plus evaluation results.}
        We report success rates (\%) under the original setting and seven
        perturbation dimensions~\cite{robotwin2-plus}. The best result in each column is shown in
        \textbf{bold}, while the second-best result is
        \underline{underlined}. 
    }
    \label{tab:libero_plus}
    \small
    \setlength{\tabcolsep}{4.2pt}
    \renewcommand{\arraystretch}{1.08}
    \begin{tabular}{lccccccccc}
        \toprule
        \textbf{Model}
        & \textbf{Original}
        & \textbf{Camera}
        & \textbf{Robot}
        & \textbf{Lang.}
        & \textbf{Light}
        & \textbf{BG}
        & \textbf{Noise}
        & \textbf{Layout}
        & \textbf{Total} \\
        \midrule

        \multicolumn{10}{c}{\textit{VLAs}} \\
        \midrule
        $\pi_0$~\citeyearpar{pi0}
            & 94.2 & 13.8 & 6.0 & 58.8 & 85.0 & 81.4 & 79.0 & 68.9 & 53.6 \\
        $\pi_0$ rerun~\citeyearpar{pi0}
            & 91.3 & 61.0 & 40.8 & 63.5 & 89.3 & 84.1 & 80.1 & 76.4 & 69.4 \\
        $\pi_0$-FAST~\citeyearpar{fasttokenizer}
            & 85.5 & 65.1 & 21.6 & 61.0 & 73.2 & 73.2 & 74.4 & 68.8 & 61.6 \\
        $\pi_{0.5}$~\citeyearpar{pi05}
            & 96.9 & 75.4 & 77.5 & 85.6 & \textbf{96.9} & \underline{94.6}
            & 89.7 & \textbf{85.7} & \underline{85.7} \\
        OpenVLA-OFT$_m$~\citeyearpar{openvlaoft}
            & 97.6 & 55.6 & 21.7 & 81.0 & 92.7 & 91.0 & 78.6 & 68.7 & 67.9 \\
        UniVLA~\citeyearpar{univla}
            & 95.2 & 1.8 & 46.2 & 69.6 & 69.0 & 81.0 & 21.2 & 31.9 & 42.9 \\
        RIPT-VLA~\citeyearpar{riptvla}
            & 97.5 & 55.2 & 31.2 & 77.6 & 88.4 & 91.6 & 73.5 & 74.2 & 68.4 \\
        X-VLA~\citeyearpar{xvla}
            & 98.1 & 23.4 & \underline{89.7} & 75.7 & 88.2 & \textbf{96.0}
            & 62.7 & 71.8 & 71.4 \\
        HoloBrain0-GD~\citeyearpar{holobrain0}
            & 96.7 & 65.5 & 58.2 & 78.7 & 88.1 & 90.3 & 66.9 & 79.5 & 74.0 \\
        ABot-M0~\citeyearpar{abotm0}
            & \textbf{98.6} & 60.4 & 67.9 & 86.4 & 96.2 & 91.6 & 86.4 & 82.6 & 80.5 \\

        \midrule
        \multicolumn{10}{c}{\textit{VLA + WM}} \\
        \midrule
        VLA-JEPA~\citeyearpar{vlajepa}
            & 97.2 & 64.2 & 67.7 & \underline{88.1} & 91.8 & 93.4
            & 65.8 & \underline{83.9} & 77.9 \\

        \midrule
        \multicolumn{10}{c}{\textit{WAMs}} \\
        \midrule
        GE-Act~\citeyearpar{geact}
            & 94.4 & 60.7 & 77.0 & 77.4 & 95.8 & 86.0 & 90.9 & 80.2 & 80.3 \\
        Cosmos Policy~\citeyearpar{cosmospolicy}
            & \underline{98.5} & \underline{75.8} & 63.3 & 81.7
            & \underline{96.5} & 88.9 & \underline{92.7} & 82.2 & 82.2 \\
        Fast-WAM~\citeyearpar{fastwam}
            & 97.6 & 16.4 & 44.5 & 68.9 & 78.2 & 53.7 & 37.7 & 60.7 & 51.5 \\

        \midrule
        \textbf{Ours}
            & 98.2
            & \textbf{60.9}
            & \textbf{96.2}
            & \textbf{91.2}
            & 86.8
            & 94.1
            & \textbf{96.4}
            & 58.4
            & \textbf{87.8} \\
        \bottomrule
    \end{tabular}
    \vspace{-0.3cm}
\end{table}

The clearest gains occur under robot-initialization, language, and observation-noise shifts. We achieve 96.2\% under robot-initialization shifts, 6.5 points above the best competitor, and rank first on language (91.2\%), respectively. Background performance is also strong at 94.1\%, although $\pi_{0.5}$ and X-VLA remain ahead. Camera shifts present a different picture: our 60.9\% success rate trails the best competitor by 14.9 points. Lighting (86.8\%) and especially layout (58.4\%) also remain challenging. These results suggest that weighting residual errors along action-responsive directions helps with some changes in embodiment and observation, but does not by itself resolve viewpoint shifts or large geometric rearrangements that require scene-level reasoning.

\subsection{RoboTwin and RoboTwin 2.0-Plus}

RoboTwin~\cite{robotwin} and RoboTwin 2.0-Plus~\cite{robotwin2-plus} stress bimanual coordination across diverse
dual-arm tasks: RoboTwin reports the Clean and Randomized settings, and
RoboTwin 2.0-Plus perturbs the same seven dimensions as LIBERO-Plus. Each
benchmark uses one unified multi-task model trained for 5 epochs and evaluated
with the same checkpoint without large-scale embodied
pretraining, which lets us read the data efficiency of ATF supervision directly.

\begin{table*}[t]
    \centering
    \caption{
        \textbf{Results on RoboTwin and RoboTwin 2.0-Plus.}
        We report success rates (\%) on RoboTwin under the Clean and Randomized settings,
        and on RoboTwin 2.0-Plus under the original setting and seven perturbation dimensions.
        For RoboTwin 2.0-Plus, all methods use a single unified model across all tasks.
        The best result in each column is highlighted in \textbf{bold}, while the
        second-best result is \underline{underlined}.
        Embodied PT.: embodied pretraining. Lang.: Language. BG: Background.
    }
    \label{tab:robotwin}

    \resizebox{\textwidth}{!}{
    \begin{tabular}{lcccc|ccccccccc}
        \toprule
        & \multicolumn{4}{c|}{\textbf{RoboTwin}}
        & \multicolumn{9}{c}{\textbf{RoboTwin 2.0-Plus}} \\
        \cmidrule(lr){2-5}
        \cmidrule(lr){6-14}

        Method
        & Embodied PT.
        & Clean
        & Rand.
        & Avg.
        & Original
        & Camera
        & Robot
        & Lang.
        & Light
        & BG
        & Noise
        & Layout
        & Total \\
        \midrule

        $\pi_{0}$~\citeyearpar{pi0}
        & $\checkmark$
        & 65.92
        & 58.40
        & 62.2
        & --
        & --
        & --
        & --
        & --
        & --
        & --
        & --
        & -- \\

        $\pi_{0.5}$~\citeyearpar{pi05}
        & $\checkmark$
        & 82.74
        & 76.76
        & 79.8
        & 78.4
        & \textbf{45.6}
        & 27.6
        & 74.4
        & 49.6
        & 71.7
        & 64.9
        & 56.8
        & 58.6 \\

        X-VLA~\citeyearpar{xvla}
        & --
        & --
        & --
        & --
        & 65.6
        & 23.2
        & 65.2
        & 64.4
        & 63.1
        & 58.6
        & 49.7
        & 34.8
        & 53.1 \\

        MOTUS~\citeyearpar{motus}
        & $\checkmark$
        & 88.66
        & 87.02
        & 87.8
        & 87.0
        & 21.6
        & \textbf{85.0}
        & 83.2
        & 84.6
        & 84.4
        & 43.1
        & 82.8
        & 71.5 \\

        LingBot-VA~\citeyearpar{lingbotva}
        & $\checkmark$
        & \textbf{92.90}
        & 91.50
        & \underline{92.2}
        & \underline{92.1}
        & 28.9
        & 36.2
        & \underline{87.3}
        & \underline{89.0}
        & \textbf{91.3}
        & \textbf{80.9}
        & \underline{87.9}
        & \underline{74.2} \\

        Fast-WAM~\citeyearpar{fastwam}
        & $\times$
        & 91.88
        & \underline{91.78}
        & 91.8
        & 91.2
        & \underline{30.4}
        & 53.2
        & 86.7
        & 88.8
        & 90.0
        & 76.4
        & 83.2
        & 72.7 \\

        \midrule

        \textbf{Ours}
        & $\times$
        & \underline{92.80}
        & \textbf{92.10}
        & \textbf{92.4}
        & \textbf{92.2}
        & 19.1
        & \underline{81.4}
        & \textbf{88.8}
        & \textbf{91.4}
        & \underline{91.2}
        & \underline{78.8}
        & \textbf{91.3}
        & \textbf{79.3} \\

        \bottomrule
    
    \end{tabular}
    }
    
\end{table*}

We reach 92.8\% on Clean and 92.1\% on Randomized, the best average of
\textbf{92.4\%}: within 0.1 points of pretrained LingBot-VA on Clean, ahead of it
by 0.6 on Randomized. The 0.7-point
Clean--Randomized gap indicates little reliance on training-scene appearance, so
ATF supervision matches WAMs that use substantial embodied pretraining from
task-specific data alone.

Under the perturbation-wise protocol we obtain the best original-setting score
(92.2\%) and raise the aggregate to \textbf{79.3\%}, 5.1 points above
LingBot-VA. We rank first under language (88.8\%), lighting (91.4\%), and layout
(91.3\%), ahead of the respective runners-up by 1.5, 2.4, and 3.4 points, and
reach 81.4\% under robot perturbations, second only to MOTUS. Gains are not
uniform: background and observation noise trail the best results by 0.1 and 2.1
points, and camera perturbation is the clearest weakness at 19.1\%. Together
with the LIBERO-Plus layout result, this indicates that a large change in
observation geometry moves the input outside the local action--world
neighborhood the tangent estimate covers, while within that neighborhood the
objective follows action-induced rather than merely visually salient variation.

\subsection{Real-World Experiments across Embodiments}
\label{sec:real_world_experiments}

\begin{figure}[h]
    \centering
    \includegraphics[width=\linewidth]{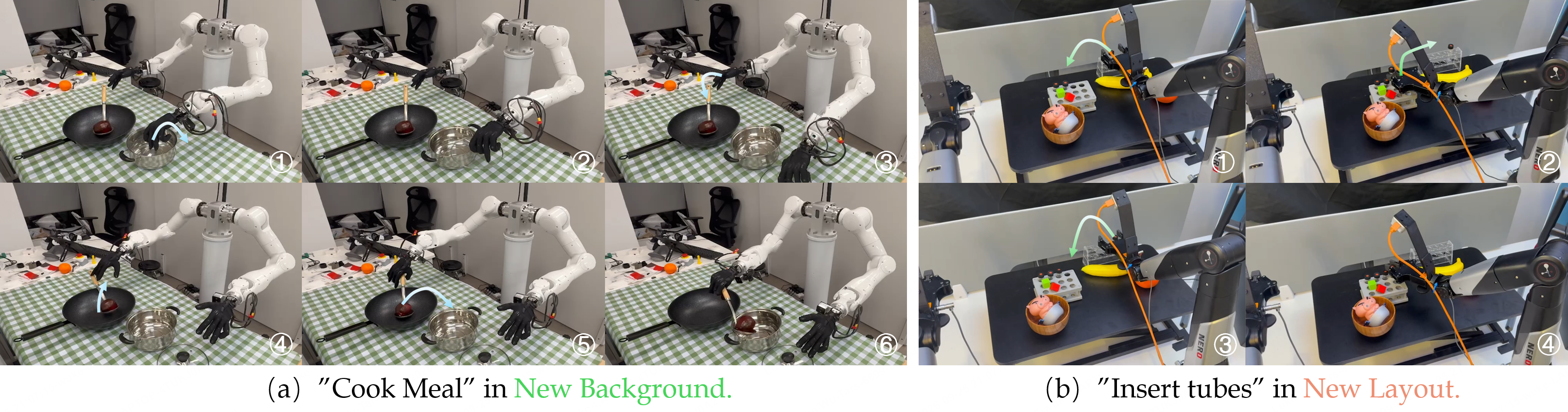}
    \caption{\textbf{Real-world rollouts across embodiments.} Left: Tianji performs bimanual cooking with dexterous hands under a new background. Right: NERO inserts test tubes with a parallel gripper under new objects layout, numbered frames show the progression of each task.}
    \label{Nero_real_world_test}
    
\end{figure}

We evaluate on two real-world platforms, NERO and Tianji, covering parallel-gripper and dexterous-hand manipulation over five tasks: test-tube insertion, bowl stacking, language-conditioned ordered object transfer, whiteboard wiping, and bimanual cooking. Each policy is tested under four conditions (standard, red light, green light, and tabletop-background replacement), with three trials per task--condition pair, we report the average progress score. Models are trained and evaluated separately per task for 10 epochs. Task descriptions and platform setups are in Appendix~\ref{sec:real_world_experiments_details}.

\begin{table*}[t]
\vspace{-0.3cm}
\centering
\caption{\textbf{Real-world evaluation on NERO and Tianji}. Scores are averaged over three trials for each task and condition. The Average column is the mean across the four conditions. }
\label{tab:real_world_evaluation}
\small
\begin{tabular*}{\textwidth}{@{\extracolsep{\fill}}llccccc}
\toprule
\multicolumn{7}{c}{\textit{7-DoF NERO Arm \& Gripper}} \\
\midrule
Task & Method & Standard & Layout 1 & Layout 2
& Red \& Green Light & Average \\
\midrule
Stack Bowls
& $\pi_{0.5}$~\citeyearpar{pi05}              & 0.33 & \underline{0.33} & \underline{0.17} & \textbf{0.50} & \underline{0.33} \\
& FastWAM~\citeyearpar{fastwam}            & \underline{0.50} & \textbf{0.50} & \underline{0.17} & \underline{0.17} & \underline{0.33} \\
& $\Delta$WAM (Ours) & \textbf{0.67} & \textbf{0.50} & \textbf{0.50} & \textbf{0.50} & \textbf{0.54} \\
\midrule
Insert Tubes
& $\pi_{0.5}$~\citeyearpar{pi05}              & \underline{0.83} & \underline{0.50} & \underline{0.50} & \underline{0.50} & \underline{0.58} \\
& FastWAM~\citeyearpar{fastwam}            & 0.50 & \underline{0.50} & 0.33 & \underline{0.50} & 0.46 \\
& $\Delta$WAM (Ours) & \textbf{1.00} & \textbf{0.83} & \textbf{0.67} & \textbf{0.67} & \textbf{0.79} \\
\midrule
\multicolumn{7}{c}{\textit{7-DoF Tianji Arm \& 20-DoF Wuji Dexterous Hand}} \\
\midrule
Task & Method & Standard & Red Light & Green Light
& New Background & Average \\
\midrule
 Move Ingredients
& $\pi_{0.5}$~\citeyearpar{pi05}               & \textbf{0.73} & \underline{0.53} & \underline{0.10} & \underline{0.10} & \underline{0.37} \\
& FastWAM~\citeyearpar{fastwam}            & 0.40 & 0.20 & \underline{0.10} & \underline{0.10} & 0.20 \\
& $\Delta$WAM (Ours) & \underline{0.63} & \textbf{0.73} & \textbf{0.60} & \textbf{0.60} & \textbf{0.64} \\
\midrule
Wipe Board
& $\pi_{0.5}$~\citeyearpar{pi05}              & \underline{0.57} & \underline{0.50} & 0.27 & 0.10 & 0.36 \\
& FastWAM~\citeyearpar{fastwam}            & 0.43 & 0.37 & \textbf{0.47} & \textbf{0.30} & \underline{0.39} \\
& $\Delta$WAM (Ours) & \textbf{0.80} & \textbf{0.53} & \underline{0.33} & \underline{0.20} & \textbf{0.47} \\
\midrule
Cook Meal
& $\pi_{0.5}$~\citeyearpar{pi05}              & \underline{0.08} & \underline{0.00} & \underline{0.00} & \underline{0.00} & \underline{0.02} \\
& FastWAM~\citeyearpar{fastwam}            & -- & -- & -- & -- & -- \\
& $\Delta$WAM (Ours) & \textbf{0.42} & \textbf{0.42} & \textbf{0.25} & \textbf{0.42} & \textbf{0.38} \\
\bottomrule
\end{tabular*}

\end{table*}

Table~\ref{tab:real_world_evaluation} compares $\pi_{0.5}$, Fast-WAM, and our
method under identical task-specific data and evaluation conditions. The NERO
tasks require precise insertion and rigid-object placement with parallel
grippers, whereas the Tianji tasks require language-conditioned sequencing,
sustained contact, bimanual coordination, and tool use with dexterous hands. ATF
supervision transfers across both regimes: on NERO, $\Delta$WAM averages 0.79 on
test-tube insertion against 0.58 for $\pi_{0.5}$ and 0.54 on bowl stacking
against 0.33 for both baselines, on Tianji it leads every task average, with the
largest margin on bimanual cooking, where the baselines score at most 0.02.

\subsection{Ablation Study}

\begin{figure*}[t]
    \vspace{-0.2cm}
    \centering
    \includegraphics[width=0.9\textwidth]{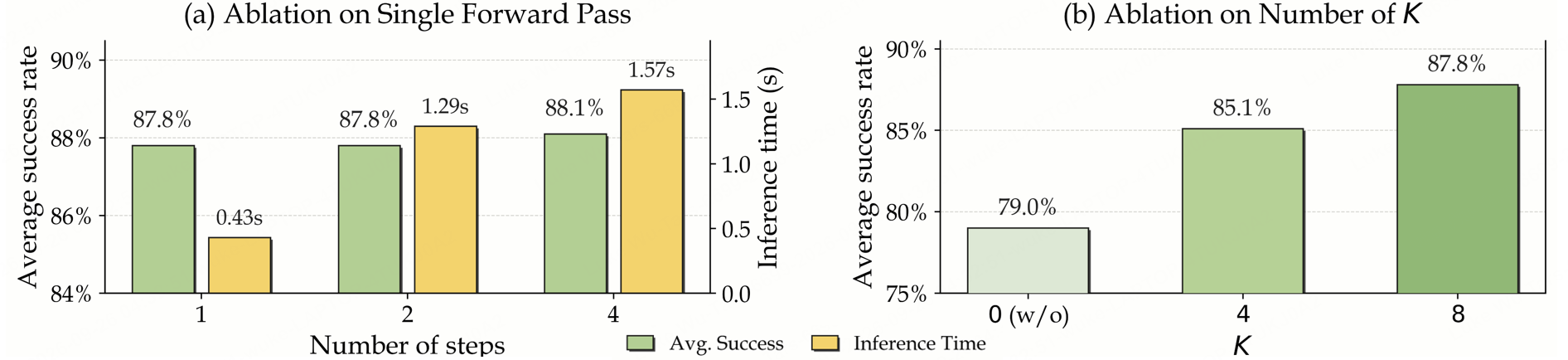}
    \vspace{-0.3cm}
    \caption{\textbf{Ablations on LIBERO-Plus.} (a) Average success and inference time for different VideoDiT denoising steps. (b) Average success for $K=0,4,8$ local action perturbations.}
    \vspace{-0.4cm}
    \label{fig:ablation}
\end{figure*}

We ablate two design choices on LIBERO-Plus: the number of VideoDiT denoising
steps at inference and the number $K$ of local action perturbations used to
construct the ATF. Figure~\ref{fig:ablation} reports average success, and panel
(a) also reports inference time.

\paragraph{VideoDiT denoising steps.}
One and two steps both achieve 87.8\% average success and four steps reach 88.1\%, while inference time rises from 0.43\,s to 1.29\,s and 1.57\,s. Three extra steps thus buy 0.3 points at 1.14\,s of additional latency, so we use one step, which keeps nearly the same success rate without iterative refinement.

\paragraph{Local action neighborhood.}
At $K=0$, the policy retains the real action and residual targets but receives no ATF loss, achieving 79.0\% success. Success rises to 85.1\% at $K=4$ and 87.8\% at $K=8$, an 8.8-point gain that motivates our use of $K=8$ throughout.

\section{Conclusion}
$\Delta$WAM learns future video dynamics and how they vary within a local neighborhood of the demonstrated action. Residual-VAE prediction keeps the full future latent while making temporal change explicit, and the ATF loss upweights errors along directions that respond to nearby actions, keeping the real future as the target. One VideoDiT pass then suffices at inference. Gains hold on LIBERO-Plus, RoboTwin, and RoboTwin 2.0-Plus, though large camera and layout changes remain challenging.

\newpage
\bibliography{main}
\bibliographystyle{iclr2027_conference}

\end{document}